\documentclass[letterpaper]{article} 
\usepackage{aaai19}  
\usepackage{times}  
\usepackage{helvet}  
\usepackage{courier}  
\usepackage{url}  
\usepackage{graphicx}  
\usepackage{ifthen}

\makeatletter
\renewcommand\@cite[2]{%
Ref.~#1\ifthenelse{\boolean{@tempswa}}
{, \nolinebreak[3] #2}{}
}
\renewcommand\@biblabel[1]{#1.}
\makeatother

\usepackage{amsmath}
\usepackage{amsfonts}

\title{Predicting Periodicity with Temporal Difference Learning}
\author{Kristopher De Asis, Brendan Bennett, Richard S. Sutton\\
Reinforcement Learning and Artificial Intelligence Laboratory, University of Alberta \\
\{kldeasis, babennet, rsutton\}@ualberta.ca
}

\begin{document}
\maketitle
\begin{abstract}
Temporal difference (TD) learning is an important approach in reinforcement learning, as it combines ideas from dynamic programming and Monte Carlo methods in a way that allows for online and incremental model-free learning. A key idea of TD learning is that it is learning predictive knowledge about the environment in the form of value functions, from which it can derive its behavior to address long-term sequential decision making problems. The agent's horizon of interest, that is, how immediate or long-term a TD learning agent predicts into the future, is adjusted through a discount rate parameter. In this paper, we introduce an alternative view on the discount rate, with insight from digital signal processing, to include complex-valued discounting. Our results show that setting the discount rate to appropriately chosen complex numbers allows for online and incremental estimation of the Discrete Fourier Transform (DFT) of a signal of interest with TD learning. We thereby extend the types of knowledge representable by value functions, which we show are particularly useful for identifying periodic effects in the reward sequence.
\end{abstract}

\section{Temporal Difference Learning}
\label{sec:tdintro}

\textit{Temporal-difference} (TD) methods~\cite{sutton1988} are an important approach in \textit{reinforcement learning} as they combine ideas from dynamic programming and Monte Carlo methods. TD allows learning to occur from raw experience in the absence of a model of the environment's dynamics, like with Monte Carlo methods, while computing estimates which bootstrap from other estimates, like with dynamic programming. This provides a way for an agent to learn online and incrementally in both long-term prediction and sequential decision-making problems.

A key view of TD learning is that it is learning testable, predictive knowledge of the environment~\cite{horde2011}. The learned value functions represent answers to predictive questions about how a signal will accumulate over time, given a way of behaving in the environment. A TD learning agent can continually compare its predictions to the actual outcomes, and incrementally adjust its world knowledge accordingly. In control problems, this signal is the reward sequence, and the value function represents the long-term cumulative reward an agent expects to receive when behaving greedily with respect to its current predictions about this signal.

A TD learning agent's time horizon of interest, or how long-term it is to predict into the future, is specified through a \textit{discount rate}~\cite{rlbook2018}. This parameter adjusts the weighting given to later outcomes in the sum of a sequence over time, trading off between only considering immediate or near-term outcomes and estimating the sum of arbitrarily long sequences. From this interpretation of its purpose, along with convergence considerations, the discount rate is restricted to be $\gamma \in [0, 1]$ in episodic problems, and $\gamma \in [0, 1)$ in continuing problems.

In this paper, we investigate whether meaningful information can be learned from relaxing the range of values the discount rate can be set to. In particular, we allow it to take on complex values, and instead restrict the magnitude of the discount rate, $\lvert \gamma \rvert$, to fall within the aforementioned ranges.

\section{One-step TD and the MDP Formalism}
\label{sec:mdps}
The sequential decision-making problem in reinforcement learning is often modeled as a \textit{Markov decision process} (MDP). Under the MDP framework, an \textit{agent} interacts with an environment over a sequence of discrete time steps. At each time step $t$, the agent receives information about the environment's current \textit{state}, $S_t \in \mathcal{S}$, where $\mathcal{S}$ is the set of all possible states in the MDP. The agent is to use this state information to select an \textit{action}, $A_t \in \mathcal{A}(S_t)$, where $\mathcal{A}(s)$ is the set of possible actions in state $s$. Based on the environment's current state and the agent's selected action, the agent receives a \textit{reward}, $R_{t+1} \in \mathbb{R}$, and gets information about the environment's next state, $S_{t+1} \in \mathcal{S}$, according to the \textit{environment model}:
$p(r,s'|s,a)={P(R_{t+1}=r,S_{t+1}=s'|S_t=s,A_t=a)}$.

The agent selects actions according to a \textit{policy}, $\pi(s,a) = P(A_t=a|S_t=s)$, which gives a probability distribution across actions $a \in \mathcal{A}(s)$ for a given state $s$, and is interested in the expected discounted return:
\begin{equation}
G_t = R_{t+1} + \gamma R_{t+2} + \gamma^2 R_{t+3} + ... = \sum_{k=0}^{T-t - 1} \gamma^{k} R_{t+k+1}
\label{eqn:returndef}
\end{equation}
given a discount rate $\gamma \in [0,1]$ and $T$ equal to the final time step in an episodic task, or $\gamma \in [0,1)$ and $T$ equal to infinity for a continuing task.

\textit{Value-based methods} approach the sequential decision-making problem by computing \textit{value functions}, which provide estimates of what the return will be from a particular state onwards. In prediction problems, also referred to as \textit{policy evaluation}, the goal is to estimate the return under a particular policy as accurately as possible, and a \textit{state-value function} is often estimated. It is defined to be the expected return when starting in state $s$ and following policy $\pi$:
\begin{equation}
v_\pi(s) = \mathbb{E}_\pi[G_t|S_t=s]
\label{eqn:vdef}
\end{equation}
For control problems, the policy which maximizes the expected return is to be learned, and an \textit{action-value function} from which a policy can be derived is instead estimated. It is defined to be the expected return when taking action $a$ in state $s$, and following policy $\pi$:
\begin{equation}
q_\pi(s,a) = \mathbb{E}_\pi[G_t|S_t=s,A_t=a]
\label{eqn:qdef}
\end{equation}
Of note, the action-value function can still be used for prediction problems, and the state-value can be computed as an expectation across action-values under the policy $\pi$ for a given state:
\begin{equation}
v_\pi(s) = \mathbb{E}_{\pi}[q_\pi(s,\cdot)] = \sum_{a}{\pi(s, a)q_\pi(s,a)}
\label{eqn:qtov}
\end{equation}
TD methods learn an approximate value function, such as $V \approx v_\pi$ for state-values, by computing an estimate of the return, $\hat{G}_t$. First, Equation \ref{eqn:qdef} can be written in terms of its successor state-action pairs, also known as the \textit{Bellman equation} for $v_\pi$:
\begin{equation}
v_\pi(s) = \sum_{a}{\pi(s,a)\sum_{r, s'}{p(r,s'|s,a)\Big(r + \gamma v_\pi(s')\Big)}}
\label{eqn:vbellman}
\end{equation}
Based on Equation \ref{eqn:vbellman}, one-step TD methods estimate the return by taking an action in the environment according to a policy, sampling the immediate reward, and bootstrapping off of the current estimates in the value function for the remainder of the return. The difference between this \textit{TD target} and the value of the previous state is then computed, and is often referred to as the \textit{TD error}. The previous state's value is then updated by taking a step proportional to the TD error with step size $\alpha \in (0, 1]$:
\begin{align}
\hat{G}_t &= R_{t+1} + \gamma V(S_{t+1})
\label{eqn:tdtarget} \\
V(S_t) &\leftarrow V(S_t) + \alpha [\hat{G}_t - V(S_t)]
\label{eqn:tdupdate}
\end{align}
Since the rewards received depend on the actions selected, the above updates will learn the expected return under the policy that is generating its behavior, and is referred to as \textit{on-policy} learning. \textit{off-policy} learning allows an agent to learn about the expected return given a policy different from the one generating an agent's behavior. One way of achieving this is through \textit{importance sampling}~\cite{pdis2000}, where with a behavior policy $\mu$ and a target policy $\pi$, an alternative update to Equation \ref{eqn:tdupdate} is:
\begin{equation}
V(S_t) \leftarrow V(S_t) + \alpha \frac{\pi(S_t, A_t)}{\mu(S_t, A_t)} [\hat{G}_t - V(S_t)]
\label{eqn:offtdupdate}
\end{equation}
This strictly generalizes the on-policy case, as the importance sampling ratio is $1$ when the two policies are identical.

\section{Complex Discounting}
\label{sec:complexdiscounting}

The discount rate has an interpretation of specifying the horizon of interest for the return, trading off between focusing on immediate rewards and considering the sum of longer sequences of rewards. It can also be interpreted as a \textit{soft termination} of the return~\cite{horde2011,tdmodel1995,nexting2014}, where an agent includes the next reward with probability $\gamma$, and terminates with probability $1 - \gamma$, receiving a terminal reward of $0$. From these interpretations, it is intuitive for the discount rate to fall in the range of $\gamma \in [0, 1)$ with the exception of episodic problems, where $\gamma$ can be equal to 1.

With considerations for convergence, assuming the rewards are bounded, restricting the discount rate to be in this range makes the infinite sum (in the continuing case) of Equation \ref{eqn:returndef} finite. However, this sum will remain finite when the magnitude of the discount rate is restricted to be $\lvert \gamma \rvert \in [0, 1)$, allowing for the use of negative discount rates up to $-1$, as well as complex discount rates within the complex unit circle.

While the use of alternative discount rates may result in some corresponding value function, a question arises regarding whether these values are meaningful, or if there is any situation in which an agent would benefit from this knowledge. First, we consider the implications of exponentiating a complex discount rate. We look at the exponential form of a complex number with unit magnitude, and note that it can be expressed as a sum of sinusoids by Euler's Formula:
\begin{equation}
e^{-i \omega}  = \cos(\omega) - i \sin(\omega)
\label{eqn:eulersformula}
\end{equation}
From this, it is evident that exponentiating a complex number to the power of $n$ corresponds to taking $n$ steps around the complex unit circle with an angle of $\omega$:
\begin{equation}
e^{-i \omega n}  = \cos(n \omega) - i \sin(n \omega)
\label{eqn:expcomplex}
\end{equation}
Using the above as a discount rate, assuming an episodic setting as it has a magnitude of $1$, we would get the following return for some angle $\omega$:
\begin{equation}
G_t^{\omega} = \sum_{k=0}^{T-t - 1} e^{-i \omega k} R_{t+k+1}
\label{eqn:complexreturn}
\end{equation}
Instead of weighting the reward sequence in a way that decays the importance of future rewards, complex discount rates weight the sequence with two sinusoids, one along the real axis and one along the imaginary axis. This can be interpreted as checking the cross correlation between a reward sequence and a sinusoid oscillating with a frequency of $\omega \frac{\textrm{rad}}{\textrm{step}}$, and effectively allows a TD learning agent to identify periodicity in the reward sequence at specified frequencies online and incrementally.

\section{The Discrete Fourier Transform}
\label{sec:dft}
The ability to identify periodicity in the reward sequence by weighting it with exponentiated complex numbers can be viewed as performing the Discrete Fourier Transform (DFT) from digital signal processing literature~\cite{fftbook1988}. The DFT is defined as follows:
\begin{equation}
X_k = \sum_{n=0}^{N-1}{x_n e^{-i 2 \pi \frac{k}{N} n}}
\label{eqn:dftdef}
\end{equation}
where $N$ is the length of the sequence, and $k$ is set to each whole number less than $N$. This can be viewed as testing whether a frequency of $2 \pi \frac{k}{N}$ exists in the sequence, for equally spaced values of $k$. If a frequency of $2 \pi \frac{k}{N}$ exists, this sum will tend to have a larger magnitude; if no such frequency exists, the terms in the sum will tend to cancel out, and $X_{k}$ will hover around zero. Acknowledging that $k$ is less than $N$, the $\frac{k}{N}$ term is in the range $[0, 1)$, and can be rewritten where the frequency is specified directly:
\begin{equation}
X_{\omega} = \sum_{n=0}^{N-1}{x_n e^{-i \omega n}}
\label{eqn:dftfreqdef}
\end{equation}
where $\omega \in [0, 2 \pi)$. This is exactly equivalent to the DFT when the length of the sequence is known and particular frequencies $\omega$ are chosen, but has similar functionality and interpretation for other sets of frequencies.

The DFT corresponds to the discrete form of the coefficients of a Fourier series, and thus each complex number encodes the amplitude and phase of a particular sinusoidal component of a sequence. Specifically, the normalized magnitude $\lvert X_{\omega} \rvert / N$ corresponds to the amplitude, and the angle between the imaginary and real components, $\angle X_{\omega} = \arctan{\frac{\textrm{Im}(X_{\omega})}{\textrm{Re}(X_{\omega})}}$, gives the phase. The DFT is also an invertible, linear transformation~\cite{fftbook1988}. With knowledge of the length of the sequence, $N$, and the sampling frequency (in Hz), denoted $f_s$, one way of reconstructing the original sequence is by computing this sum of sinusoids:

\begin{equation}
x_n = \sum_{k=0}^{N-1}{\frac{\lvert X_{k} \rvert}{N} \cos\bigg(2\pi\frac{k}{N}f_s n + \angle X_k\bigg)}
\label{eqn:sumsines}
\end{equation}

In the context of a TD learning agent with a complex discount rate, the learned approximate values can be seen as computing the \textit{expected DFT} of the reward sequence from a specified state onwards, and allows for extraction of the corresponding amplitude and phase information. However, the expected length of the sequence is typically not known by the agent, resulting in unnormalized amplitude information.

\section{Revisiting Continuing Problems}
\label{sec:continuingprobs}

The DFT is computed with complex numbers that have a magnitude of $1$, and a discount rate of $\gamma=1$ (corresponding to $\omega=0$) would only work in the episodic setting. To see the effects of using a complex discount rate with a magnitude less than $1$, we introduce an amplitude parameter $A \in [0, 1)$ to the exponential form of a complex number:
\begin{equation}
\gamma = Ae^{-i \omega}
\label{eqn:continuinggamma}
\end{equation}
which results in a complex number with a magnitude of $A$. Substituting this in the summation in Equation \ref{eqn:returndef} gives:
\begin{equation}
G_t^{\omega} = \sum_{k=0}^{T-t - 1} e^{-i \omega k} A^k R_{t+k+1}
\label{eqn:continuingreturn}
\end{equation}
which can be seen as computing the DFT of a real discounted return with a discount rate of $A$. That is, a TD learning agent would still learn an expected DFT, but of the reward sequence over an exponentially decaying window determined by $A$. While discounting can distort the signal, it primarily affects the low frequencies which are unable to complete an oscillation within a discount rate's effective horizon.

\section{Existence and Uniqueness of Value Function}
\label{sec:vexists}

Perhaps surprisingly, the value function is well-defined for complex-valued discounting when $|\gamma| < 1$, similar to the more familiar case with a real-valued discount factor.

We first illustrate this in the continuing setting (i.e., the MDP has no terminal states).
Consider the aperiodic, irreducible Markov chain with state transition matrix $P \in \mathbb{R}^{N \times N}$, and expected reward vector $r$ with entries $\mathbb{E}[R_{t+1} | S_{t} = s]$.
Such a transition matrix has eigenvalues $1 = \lambda_{1} > \lambda_{2} \geq \cdots \geq \lambda_{N} > 0$ ~\cite{rosenthal1995convergence}.
The vector of expected returns after $n$ transitions is: 

\begin{equation}
	v^{n} = \sum_{k=0}^{n-1} (\gamma P)^n r 
\end{equation}

Where $v_{i}^{n}$ represents the expected return conditioned on starting in state $i$.

For the case where $|\gamma| < 1$, we have that $\lVert \gamma P \rVert = \beta < 1$, and that the partial sums of the matrix series $A_{n} = \sum_{k=0}^{n-1} (\gamma P)^{k}$ satisfy (for $n > m$):

\begin{equation}
\label{eq:matrix-series-cauchy}
\begin{aligned}
  \lVert A_{n} - A_{m} \rVert
  &= \lVert (\gamma P)^{m} + \cdots + (\gamma P)^{n-1} \rVert 
  \\
  &\leq {\lVert (\gamma P) \rVert}^{m} + \cdots + {\lVert (\gamma P) \rVert}^{n-1}
  \\
  &\leq \beta^{m} + \cdots \beta^{n-1}
  \\
  &= \beta^{m} \sum_{k=0}^{n-m-1} \beta^{k} < \frac{\beta^{m}}{1 - \beta}
\end{aligned}
\end{equation}

Thus the partial sums become arbitrarily close together as $m$ and $n$ grow larger.
Stated more formally, $\{A_{n}\}_{n}^{\infty}$ is a Cauchy sequence, and therefore convergent.
We can then observe that:

$$
\begin{aligned}
  (I - \gamma P) \sum_{k=0}^{n} \gamma^{k} P^{k}
  &= I + \gamma P - \gamma P + \cdots - (\gamma P^{n+1})
  \\
  &= I - (\gamma P)^{n+1}
\end{aligned}
$$

Taking the limit, we note

$$
\begin{aligned}
	\lim_{n \rightarrow \infty} (I - \gamma P) \sum_{k=0}^{n} \gamma^{k} P^{k}
    &=
    I - \lim_{n \rightarrow \infty} (\gamma P)^{n+1} 
    \\
    (I - \gamma P) \sum_{k=0}^{\infty} \gamma^{k} P^{k}
    &= I
    \\
    \Rightarrow (I - \gamma P)^{-1} 
    &= \sum_{k=0}^{\infty} \gamma^{k} P^{k}
\end{aligned}
$$

The preceding facts come together to show that the limit of the matrix series exists and is equal to $(I - \gamma P)^{-1}$.
Thus we have: 

\begin{equation}
  \begin{aligned}
	v &= \lim_{n \rightarrow \infty} v^{n}  = \sum_{k=0}^{\infty} (\gamma P)^{k} r
    \\
    &= (I - \gamma P)^{-1} r
  \end{aligned}
\end{equation}

As in the usual setting with $\gamma \in [0, 1)$.

For the episodic setting (i.e., where $P$ contains some absorbing states) we note that that we have $\lVert P \rVert = \beta < 1$, assuming that $P$ is aperiodic and indecomposable.
This implies that the previous argument for the convergence of the matrix series (in \ref{eq:matrix-series-cauchy}) holds, and that $v = (I - \gamma P)^{-1} r$ as before.

Therefore, the value function is well-defined for complex discount factors with $|\gamma| < 1$ (or $|\gamma| \leq 1$ for the episodic setting), in the sense that it exists and is unique.


\section{Experiments}

In this section, we detail several experiments involving TD learning agents using complex-valued discounting.

\label{sec:experiments}

\subsection{Checkered Grid World}

The \textit{checkered grid world} environment consists of a 5 $\times$ 5 grid of states with terminal states in the top-left and bottom-right corners. The actions consist of deterministic 4-directional movement, and moving off of the grid transitions the agent to the same state. The agent starts in the center, and the board is colored with a checkered pattern with colors representing the reward distribution. Transitioning into a white cell results in a reward of 1, transitioning into a gray cell results in a reward of -1, and transitioning to a terminal state ends the episode with a reward of 11. A diagram of the environment can be seen in Figure \ref{fig:cgwenv}. This pattern introduces an alternating pattern of 1 and -1 in the reward sequence. Given the interpretation of complex discounting as computing the DFT, we would like to see whether an agent using complex discount rates can pick up on this periodic pattern. We would also like to qualitatively assess how well the expected reward sequence can be reconstructed through Equation \ref{eqn:sumsines} (given knowledge of the expected sequence length).

\begin{figure}[h!]
	\centering
	\includegraphics[width=0.6\linewidth]{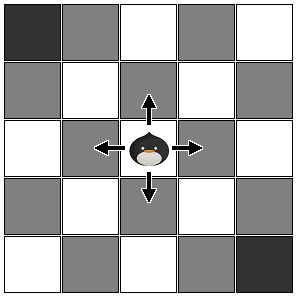}
	\caption[The checkered grid world environment]{The checkered grid world environment. Transitioning into to a white square results in a reward of 1, and transitioning into a gray square results in a reward of -1. The top left and bottom right corners represent terminal states where transitioning into them results in a reward of 11. It was set up as an on-policy policy evaluation task with an agent behaving under an equiprobable-random behavior policy.}
	\label{fig:cgwenv}
\end{figure}

This environment was treated as an on-policy policy evaluation task with no discounting ($\lvert\gamma\rvert = 1$). The agent behaved under an equiprobable-random behavior policy, which results in an expected episode length of 37.33 steps. Because the reconstruction of the reward sequence requires an integer sequence length, we round this up to 38 steps. The agent learned 114 (a multiple of 38) value functions in parallel corresponding to equally spaced frequencies in the range $\omega \in [0, 2\pi)$. Action-values were learned using the Expected Sarsa algorithm~\cite{expsarsa2009}, and state-values were computed from the learned action-values through Equation \ref{eqn:qtov}.

\begin{figure}[h!]
	\centering
	\includegraphics[width=1.0\linewidth]{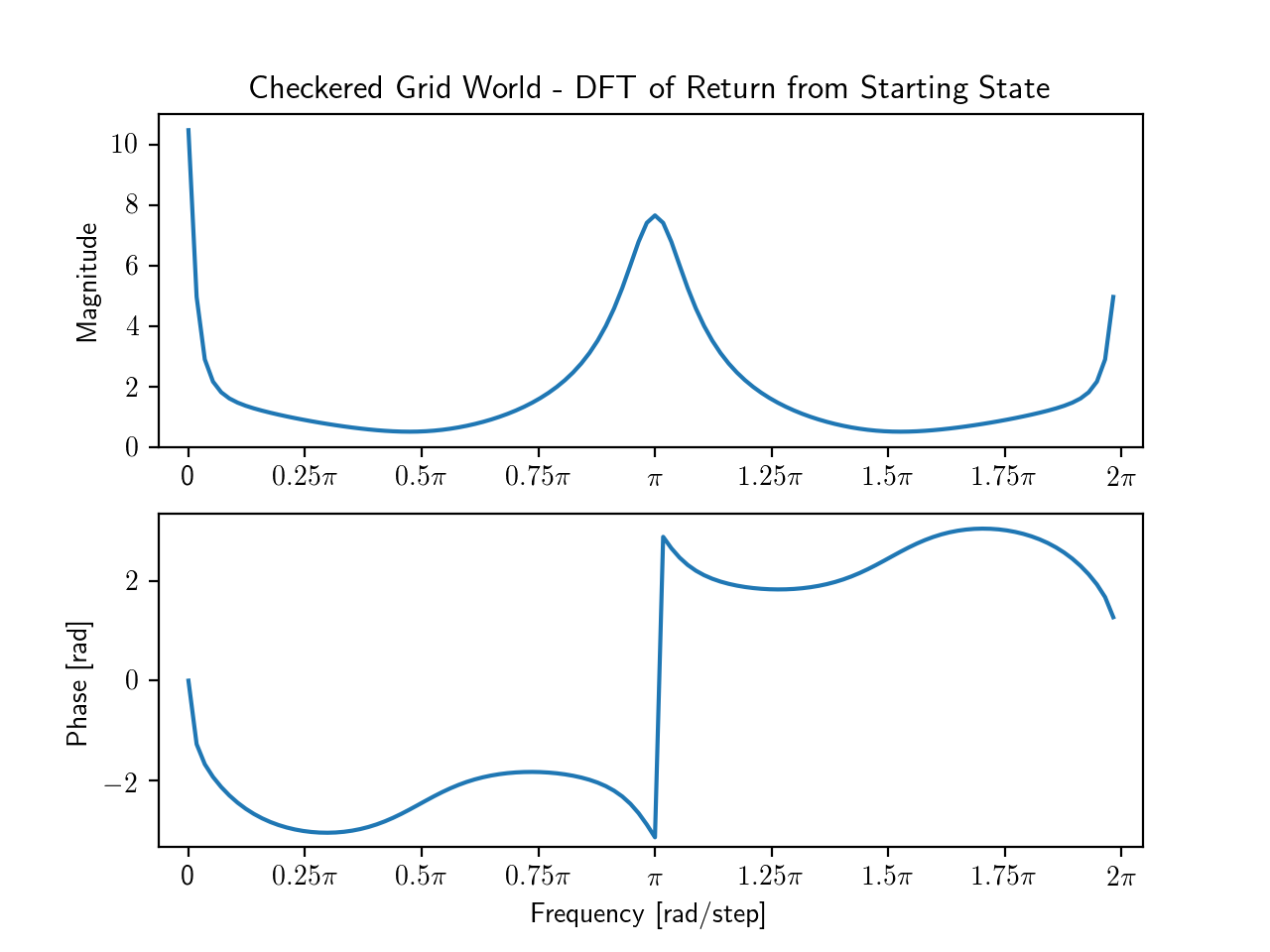}
	\caption[Checkered grid world results]{Checkered grid world policy evaluation results. The magnitude of the $\omega=0$ point corresponds to the standard undiscounted expected return under the behavior policy. The magnitudes of the other frequencies are interpreted as a measure of how confident the agent is that a frequency exists in the reward sequence. Results are averaged over 100 runs, and standard errors are less than a line width.}
	\label{fig:cgw_results}
\end{figure}

We performed 100 runs of 250 episodes, and the value of the starting state, represented by the complex number's magnitude and phase information, was plotted for each frequency after the 250th episode. The resulting learned DFT of the starting state can be seen in Figure \ref{fig:cgw_results}. Of note, the specified frequencies are normalized by the agent's sampling frequency (in Hz). Under the assumption that the agent is sampling at 1 Hz, the $\omega = 2\pi$ frequency corresponds to one sample per time step. Also, when computing the DFT of a real-valued signal, it will be symmetric about half of the sampling frequency~\cite{fftbook1988}. This ``folding'' frequency is referred to as the \textit{Nyquist frequency}, which acts as a limit for the largest detectable frequency. Frequencies larger than this would be under-sampled and subject to aliasing.

In the learned DFT, the magnitude of the value at $\omega = 0$ corresponds to the expected return with a discount rate of $\gamma = e^{-i0}=1$. That is, it is what a standard TD learning agent with a non-oscillatory discount rate would have learned. The magnitudes of the values at other frequencies are interpreted as a measure of confidence in a particular frequency existing in the reward sequence, as exact amplitude information would require normalization by sequence length. It can be seen that there is relatively large magnitude at the frequency $\omega = \pi$, which corresponds to half of the agent's sampling frequency. If the agent is sampling at a rate of 1 Hz, or 1 sample per time step, this means that it has large confidence in an oscillation at a rate of half a cycle per time step. This corresponds to the rewards alternating between 1 and -1 in the environment, as this pattern takes two time steps to complete a cycle.

\begin{figure}[h!]
	\centering
	\includegraphics[width=1.0\linewidth]{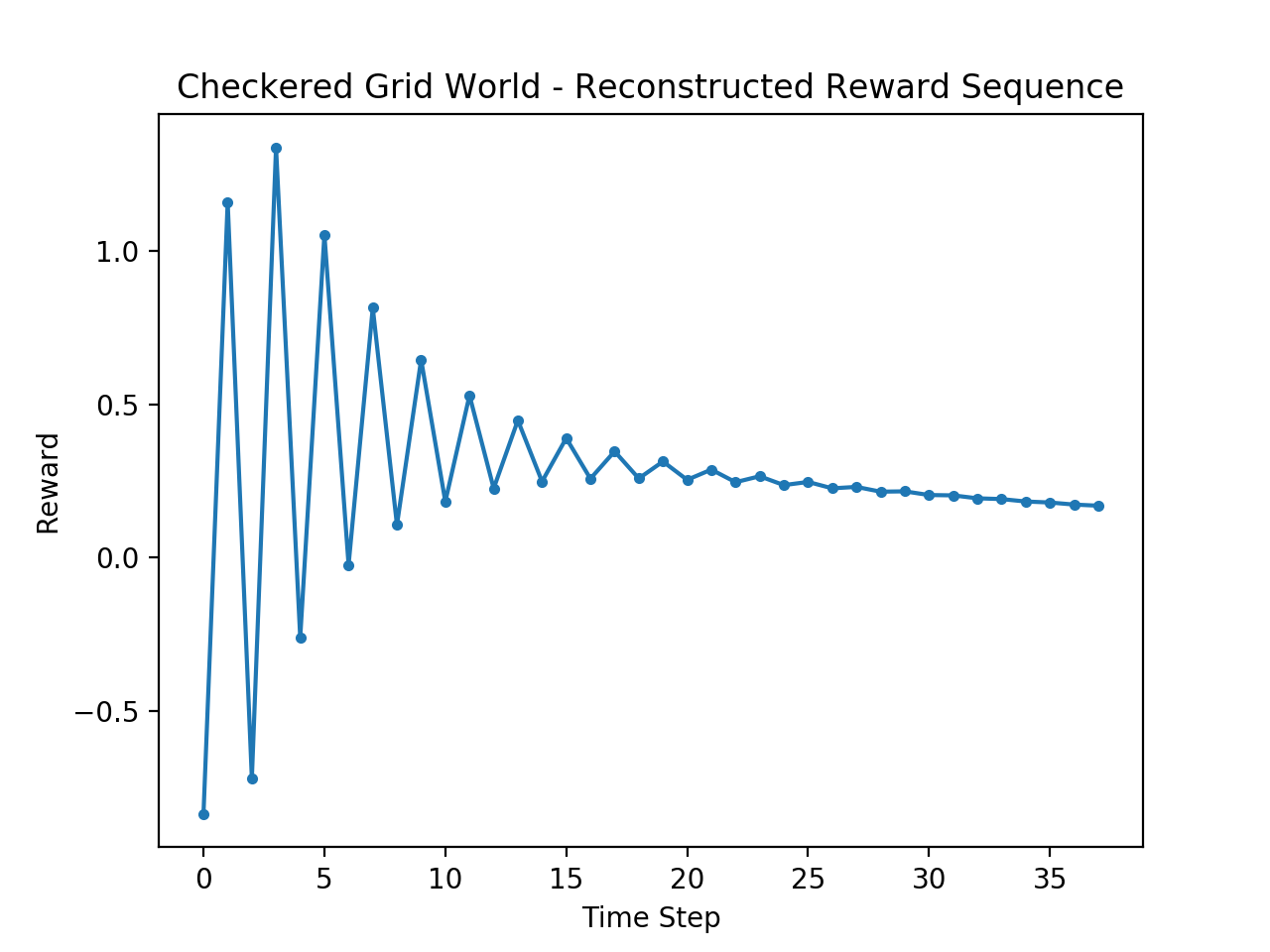}
	\caption[Checkered grid world reward sequence reconstruction]{Checkered grid world reward sequence reconstruction. It contains an apparent 0.5 Hz oscillation. The oscillation has a positive mean, corresponding to the large positive reward upon termination. The sum of this sequence is equal to the learned value of the starting state for $\omega = 0$.}
	\label{fig:cgw_recon}
\end{figure}

Next, we try to reconstruct the expected reward sequence by computing a sum of sinusoids. Using a sequence length of 38, we use the learned complex values corresponding to 38 equally spaced frequencies in $[0, 2\pi)$, and evaluate Equation \ref{eqn:sumsines} up to the 38th time step. The resulting reconstructed reward sequence can be seen in Figure \ref{fig:cgw_recon}.

One might intuitively expect the reward sequence to consist of an alternating sequence of 1 and -1, and ending with an 11. Qualitatively, the reconstructed signal does not fit this intuition, but still captures several aspects of the structure of the return. For example, the apparent oscillations in the sequence are at 0.5 Hz, and they begin at an approximate amplitude of 1. The oscillations also have a positive mean, corresponding to the large positive reward upon termination. Also, if we compute the sum of the reconstructed sequence, we get the learned value of the starting state for $\omega = 0$ (the standard undiscounted return). There are several reasons why the reconstruction wouldn't completely match the aforementioned intuition. One reason is due to cases where the agent tries to move off of the grid. Doing so transitions the agent to the same state, which may break or shift the periodic pattern each time this occurs. The earliest an agent can bump into a wall is in 3 steps, and is approximately where the exponential decay begins in the reconstructed sequence. Another reason is that the expected return consists of averaging sequences from varying episode lengths and this is an attempt at reconstructing a sequence over a fixed length (which is rounded up from the expected episode length). This would shift where in the sequence the terminal reward appears, and end up distributing it as the mean of the oscillation.

\subsection{Wavy Ring World}

The previous experiment was done in an undiscounted, episodic, tabular setting. To see whether we can achieve similar results in a continuing setting with function approximation, we designed the \textit{wavy ring world} environment. This environment consists of 20 states arranged in a ring. Each state has a single action which moves it to the next state in a fixed direction along the ring. We used tile coding~\cite{tilecoding} to produce, for each state, a binary feature vector to be used with linear function approximation. Specifically, the 20 states were covered by 6 overlapping tilings where each tile spanned 1/3-rd of the 20 states. This resulted in 6 active features for a given state, and relatively broad generalization between states. The reward for leaving a state $s \in \{0, 1, 2, ..., 19\}$, $R(s)$, consisted of the sum of four state-dependent sinusoids with periods of 2, 4, 5, and 10 states:

\begin{equation}
R(s) = \cos\bigg(\frac{2 \pi}{2}s\bigg) + \sin\bigg(\frac{2 \pi}{4}s\bigg) + \sin\bigg(\frac{2 \pi}{5}s\bigg) + \sin\bigg(\frac{2 \pi}{10}s\bigg)
\label{eqn:wgw_r}
\end{equation}

\begin{figure}[h!]
	\centering
	\includegraphics[width=0.8\linewidth]{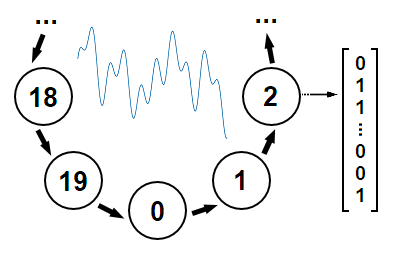}
	\caption[The wavy ring world environment]{The wavy ring world environment. Each state has one action which deterministically transitions to the next state in a fixed direction along the ring. The states produce binary feature vectors, and rewards are drawn from a sum of state-dependent sinusoids of varying frequencies.}
	\label{fig:cgwenv}
\end{figure}

\begin{figure}[h!]
	\centering
	\includegraphics[width=1.0\linewidth]{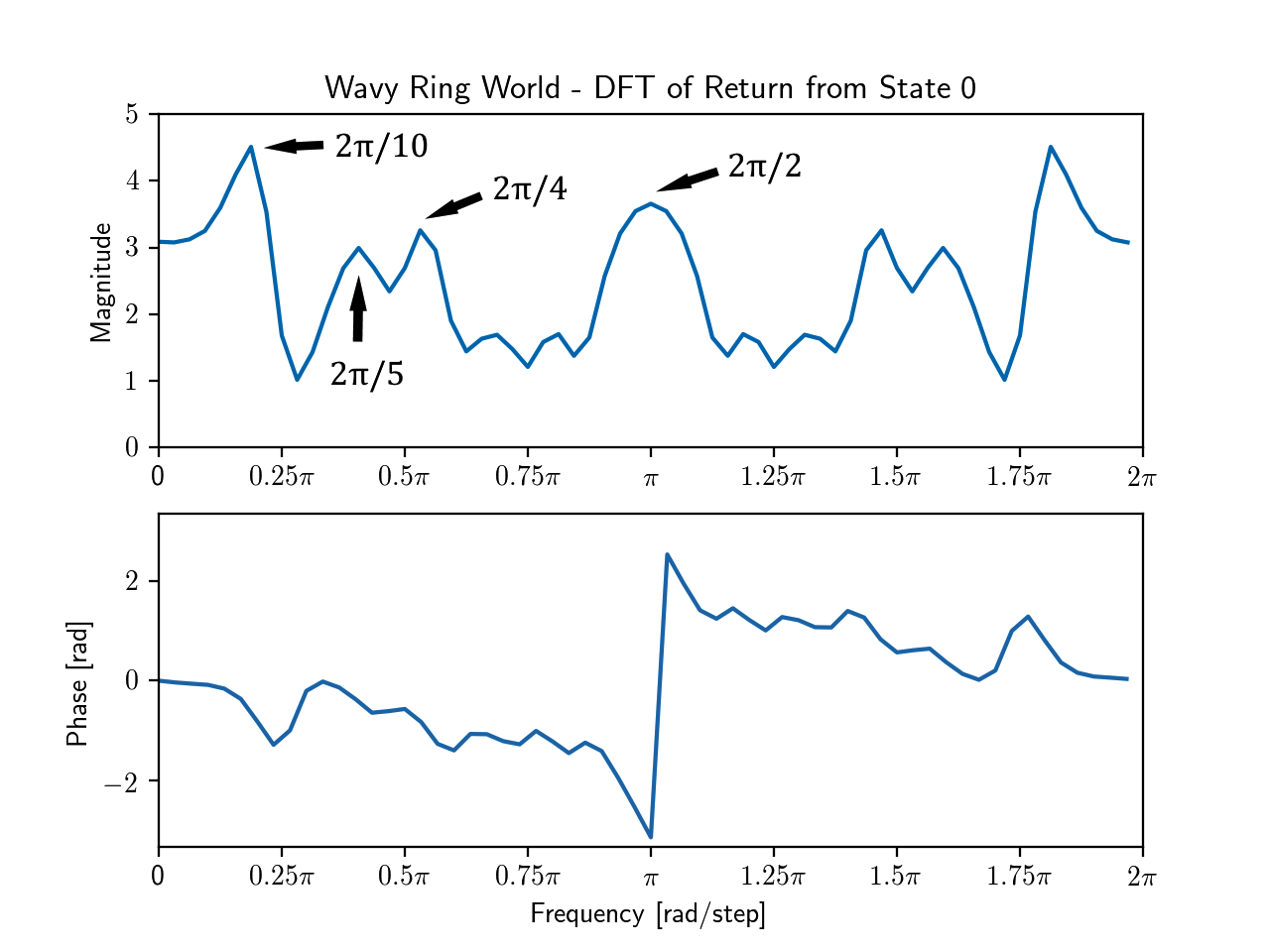}
	\caption[Wavy ring world results]{Wavy ring world results. Despite the use of discounting, as well as function approximation, the agent was still able to identify the frequencies in the reward signal.}
	\label{fig:wrw_results}
\end{figure}

A TD agent learned a set of complex-valued weights for each of 64 equally spaced frequencies in the range $\omega \in [0, 2\pi)$, and the magnitude of each discount rate was $\lvert \gamma \rvert = 0.9$. As there is no stochasticity in the transitions, we performed 1 run of 15,000 steps, with the agent starting in state 0. We extracted the state values from the learned weights, and the resulting DFT of the return from state 0 can be seen in Figure \ref{fig:wrw_results}.

In the learned DFT, we can see that despite the lower discount rate magnitude, and the use of function approximation, it still has relatively large peaks at various frequencies in the magnitude plot. Looking at the frequencies at which these peaks occur (up until the Nyquist frequency), they correspond to the frequencies of the reward function $R(s)$ in Equation \ref{eqn:wgw_r}.

\section{Discussion}
\label{sec:discussion}

From our experiments, we showed that a TD agent using complex discount rates can identify periodic patterns in the return. This is due to complex discount rates being closely related to the DFT, which a TD learning agent can be seen as incrementally estimating. We also showcased a simple way of inverting the DFT, using knowledge of the sequence length, in an attempt to reconstruct the original reward sequence. This reconstructed reward sequence contained several features pertaining to the structure of the reward sequence: An oscillation at a particular frequency and amplitude, a positive average reward, and a sum equal to the standard expected return.

Our experiments focused on a case where the periodicity came from the environment. This may have implications for reinforcement learning approaches for problems with sound or image data, as the DFT is typically used as an offline post-processing tool in those applications. In general, this approach would identify policy-contingent frequency information, as the expected return is computed under a particular policy. One could imagine an agent behaving under a policy which led it in circles. This would induce similar alternating behavior in the experienced reward sequence without this explicit structure in the environment. An example of an application involving policies containing cyclic behavior is robot gait training. If the rewards are set to be a robot's joint position, it would allow the robot to be aware of periodicity in tasks involving repetitive motion, such as walking. Such awareness of periodicity also has implications in the options framework~\cite{options1999}, as it may offer insight regarding where an option should terminate. It may also have use in exploration, where if state features are used as rewards, an agent actively avoiding periodicity might lead it to seek out novel states.

With the ability to invert the DFT and roughly reconstruct the expected reward sequence (given a sequence length), an agent would have access to information regarding the structure of the sequence.  This may be able to inform decisions based on properties like reward sparsity, or noise in the reward signal. Reconstructing the sequence can be seen as recovering the information lost from computing the sum of the rewards, which is different but comparable to distributional reinforcement learning~\cite{distribrl2017,quantilereg2017} which recovers the information lost from computing the expectation of this sum.

There has been prior work on using a Fourier basis as a representation for reinforcement learning problems~\cite{fourierbasis2011}. Using the learned value functions as a state representation, complex discounting may allow for incrementally estimating a similar representation. Also, in the deep reinforcement learning setting, learning about many frequencies in parallel may have the representation learning benefits of predicting many auxiliary tasks at once~\cite{unreal2016}.



\section{Conclusions}
\label{sec:conclusions}

In this paper, we showed that meaningful information can be learned by allowing the discount rate in TD learning to take on complex numbers. The learned complex value functions can be interpreted as incremental estimation of the DFT of a signal of interest. From this DFT interpretation, a complex discount rate corresponds to a particular frequency, and the magnitude of the learned complex value represents an agent's confidence in the frequency being present in the reward sequence. By learning several complex value functions in parallel, in both a tabular setting and one with function approximation, we showed that a TD learning agent was able to pick up on periodic structure in the reward sequence.

We also showed that information regarding the structure of the reward sequence is encoded in the resulting DFT. Because the DFT is invertible (with knowledge of the sequence length), we showed that an expected reward sequence can be reconstructed from the learned DFT. The resulting reconstructed sequence had qualitative properties that seemed reasonable for the given environment. It may be possible to infer the structure of the return from the phase information directly (without having to invert the DFT), but we leave that as an avenue for future research.

\section{Acknowledgments}
The authors thank Roshan Shariff for insights and discussions contributing to the results presented in this paper, and the entire Reinforcement Learning and Artificial Intelligence research group for providing the environment to nurture and support this research. We gratefully acknowledge funding from Alberta Innovates -- Technology Futures, Google Deepmind, and from the Natural Sciences and Engineering Research Council of Canada.

\bibliography{references}

\begin{thebibliography}{}

\bibitem[\protect\citeauthoryear{Bellemare, Dabney, and
  Munos}{2017}]{distribrl2017}
Bellemare, M.~G.; Dabney, W.; and Munos, R.
\newblock 2017.
\newblock A distributional perspective on reinforcement learning.
\newblock In {\em {ICML}}, volume~70 of {\em Proceedings of Machine Learning
  Research},  449--458.
\newblock {PMLR}.

\bibitem[\protect\citeauthoryear{Brigham}{1988}]{fftbook1988}
Brigham, E.~O.
\newblock 1988.
\newblock {\em The {F}ast {F}ourier {T}ransform and Its Applications}.
\newblock Upper Saddle River, NJ, USA: Prentice-Hall, Inc.

\bibitem[\protect\citeauthoryear{Dabney \bgroup et al\mbox.\egroup
  }{2017}]{quantilereg2017}
Dabney, W.; Rowland, M.; Bellemare, M.~G.; and Munos, R.
\newblock 2017.
\newblock Distributional reinforcement learning with quantile regression.
\newblock {\em CoRR} abs/1710.10044.

\bibitem[\protect\citeauthoryear{Jaderberg \bgroup et al\mbox.\egroup
  }{2016}]{unreal2016}
Jaderberg, M.; Mnih, V.; Czarnecki, W.~M.; Schaul, T.; Leibo, J.~Z.; Silver,
  D.; and Kavukcuoglu, K.
\newblock 2016.
\newblock Reinforcement learning with unsupervised auxiliary tasks.
\newblock {\em CoRR} abs/1611.05397.

\bibitem[\protect\citeauthoryear{Konidaris, Osentoski, and
  Thomas}{2011}]{fourierbasis2011}
Konidaris, G.~D.; Osentoski, S.; and Thomas, P.~S.
\newblock 2011.
\newblock Value function approximation in reinforcement learning using the
  {F}ourier basis.
\newblock In {\em Proceedings of the Twenty-Fifth Conference on Artificial
  Intelligence},  380--385.

\bibitem[\protect\citeauthoryear{Modayil, White, and
  Sutton}{2014}]{nexting2014}
Modayil, J.; White, A.; and Sutton, R.~S.
\newblock 2014.
\newblock Multi-timescale nexting in a reinforcement learning robot.
\newblock {\em Adaptive Behaviour} 22(2):146--160.

\bibitem[\protect\citeauthoryear{Precup, Sutton, and Singh}{2000}]{pdis2000}
Precup, D.; Sutton, R.~S.; and Singh, S.~P.
\newblock 2000.
\newblock Eligibility traces for off-policy policy evaluation.
\newblock In Kaufman, M., ed., {\em Proceedings of the 17th International
  Conference on Machine Learning},  759--766.

\bibitem[\protect\citeauthoryear{Rosenthal}{1995}]{rosenthal1995convergence}
Rosenthal, J.~S.
\newblock 1995.
\newblock Convergence rates for {M}arkov chains.
\newblock {\em Siam Review} 37(3):387--405.

\bibitem[\protect\citeauthoryear{Sutton and Barto}{2018}]{rlbook2018}
Sutton, R.~S., and Barto, A.~G.
\newblock 2018.
\newblock {\em Reinforcement Learning: An Introduction}.
\newblock 2nd edition.
\newblock Manuscript in preparation.

\bibitem[\protect\citeauthoryear{Sutton \bgroup et al\mbox.\egroup
  }{2011}]{horde2011}
Sutton, R.~S.; Modayil, J.; Delp, M.; Degris, T.; Pilarski, P.~M.; White, A.;
  and Precup, D.
\newblock 2011.
\newblock Horde: A scalable real-time architecture for learning knowledge from
  unsupervised sensorimotor interaction.
\newblock In {\em {AAMAS}},  761--768.
\newblock {IFAAMAS}.

\bibitem[\protect\citeauthoryear{Sutton, Precup, and Singh}{1999}]{options1999}
Sutton, R.~S.; Precup, D.; and Singh, S.~P.
\newblock 1999.
\newblock Between {MDP}s and semi-{MDP}s: A framework for temporal abstraction
  in reinforcement learning.
\newblock {\em Artificial Intelligence} 112(1-2):181--211.

\bibitem[\protect\citeauthoryear{Sutton}{1988}]{sutton1988}
Sutton, R.~S.
\newblock 1988.
\newblock Learning to predict by the methods of temporal differences.
\newblock {\em Machine learning} 3(1):9--44.

\bibitem[\protect\citeauthoryear{Sutton}{1995}]{tdmodel1995}
Sutton, R.~S.
\newblock 1995.
\newblock {TD} model: Modeling the world at a mixture of time scales.
\newblock Technical report, Amherst, MA, USA.

\bibitem[\protect\citeauthoryear{Sutton}{1996}]{tilecoding}
Sutton, R.~S.
\newblock 1996.
\newblock Generalization in reinforcement learning: Successful examples using
  sparse coarse coding.
\newblock In {\em Advances in Neural Information Processing Systems 8},
  1038--1044.
\newblock MIT Press.

\bibitem[\protect\citeauthoryear{van Seijen \bgroup et al\mbox.\egroup
  }{2009}]{expsarsa2009}
van Seijen, H.; van Hasselt, H.; Whiteson, S.; and Wiering, M.
\newblock 2009.
\newblock A theoretical and empirical analysis of {E}xpected {S}arsa.
\newblock In {\em Proceedings of the IEEE Symposium on Adaptive Dynamic
  Programming and Reinforcement Learning},  177--184.

\end{thebibliography}
\bibliographystyle{aaai}
\end{document}